\documentclass{article}
\usepackage{spconf,amsmath,graphicx}

\usepackage{enumitem}
\usepackage{booktabs}   
\usepackage{tabularx}   
\usepackage{caption}    
\usepackage{booktabs,tabularx,makecell}

\setlist{nosep, leftmargin=14pt}

\usepackage{mwe} 

\title{Robust by Design: A Continuous Monitoring and Data Integration Framework for Medical AI}
\name{
\shortstack[c]{%
Mohammad Daouk$^{1}$, Jan Ulrich Becker$^{2}$, Neeraja Kambham$^{3}$,\\ Anthony Chang$^{4}$, Chandra Mohan$^{1,*}$, Hien Nguyen$^{1,*}$}}
\address{$^{1}$University of Houston, Houston, TX, USA\\
$^{2}$University Hospital Cologne, Cologne, Germany\\
$^{3}$Stanford University, Stanford, CA, USA\\
$^{4}$The University of Chicago, Chicago, IL, USA\\
}
\begin{document}
%
\maketitle
\begingroup
\renewcommand\thefootnote{*}
\footnotetext{These authors jointly supervised this work.}
\endgroup
\begin{abstract}

Adaptive medical AI models often face performance drops in dynamic clinical environments due to \textbf{data drift}. We propose an \textbf{autonomous continuous monitoring and data integration framework} that maintains robust performance over time. Focusing on glomerular pathology image classification (proliferative vs. non-proliferative lupus nephritis), our three-stage method uses multi-metric feature analysis and Monte Carlo dropout-based uncertainty gating to decide when to retrain on new data. Only images statistically similar to the training distribution (via Euclidean, cosine, Mahalanobis metrics) and with low predictive entropy are integrated. The model is then incrementally retrained with these images under strict performance safeguards (no metric degradation >5\%). In experiments with a ResNet18 ensemble on a multi-center dataset, the framework \textbf{prevents performance degradation}: new images were added without significant change in AUC (~0.92) or accuracy (~89\%). This approach addresses data shift and avoids catastrophic forgetting, enabling \textbf{sustained learning} in medical imaging AI.

\end{abstract}
\begin{keywords}
Continuous monitoring, Data drift, Continual learning, Uncertainty quantification, Monte Carlo dropout, Medical AI, Renal Pathology, Glomerular classification, Lupus nephritis, Model robustness, Digital pathology.
\end{keywords}
\section{Introduction}

Glomerular classification in lupus nephritis is critical for treatment decisions ~\cite{ref1}. However, manual nephropathology review is labor-intensive and prone to inter-observer variability ~\cite{ref2}, and AI models can suffer performance degradation as incoming data deviate from the training distribution. Distribution shifts in patient populations or imaging protocols often cause static models to become unreliable ~\cite{ref3}. Continual learning paradigms have been proposed to update models continuously, but they risk \textbf{catastrophic forgetting} of prior knowledge ~\cite{ref4}. There is a need for strategies that allow models to learn from new data while preserving existing performance. We address this by combining drift detection, uncertainty-based filtering, and safe model updates in one framework.

\section{Literature Review}

Continual or lifelong learning methods aim to handle evolving data without retraining from scratch. Surveys highlight catastrophic forgetting and data drift as key challenges in medical AI ~\cite{ref5}. Techniques like Elastic Weight Consolidation add regularization to mitigate forgetting ~\cite{ref4}, and other strategies use memory replay or dynamic architectures. \textbf{Uncertainty quantification} has emerged as vital for model safety: Monte Carlo dropout and related techniques provide confidence estimates to flag unreliable predictions ~\cite{ref6}. Lambert et al. (2024) review how uncertainty methods improve trust in deep learning models for medical imaging ~\cite{ref6}. These measures can assist in detecting \textbf{data drift}: for instance, Choi et al. (2023) showed that explainability and uncertainty tools help identify when a radiology AI encounters distribution shifts ~\cite{ref3}. In general AI literature, concept drift research emphasizes monitoring data statistics to trigger model updates ~\cite{ref7}. Our work builds on these insights by using feature-space distances and predictive uncertainty to guide continuous learning.

\section{Methodology}

\subsection{Dataset and Problem Setup}

We address a binary classification task: distinguishing proliferative versus non-proliferative changes in glomerular images from lupus nephritis per ISN/RPS criteria. ``Proliferative'' requires any of the following: endocapillary hypercellularity, membranoproliferative pattern, fibrinoid necrosis, or crescents; ``non-proliferative'' is the absence of all four. 

The dataset includes 9,674 expert-labeled glomerular patches from University Hospital Cologne (Germany), Stanford University, and the University of Chicago, spanning brightfield stains (hematoxylin, PAS, trichrome, silver) across ISN/RPS Classes I–V (including III+V and IV+V). Glomeruli were extracted from whole-slide images (WSIs) using nephropathologist (JUB) annotations and labeled by the same expert on Labelbox. Class counts: 7,767 non-proliferative and 1,907 proliferative.

\subsection{Data Partitioning and Cross-Validation}

We split data into: Base (85\%) for stratified 5-fold cross-validation; New Images (1.5\%) for adaptive monitoring; and Test (13.5\%) held out for benchmarking and uncertainty thresholding. Stratification preserves class balance per fold. Each fold yields a model checkpoint for ensemble-based uncertainty estimation.

\subsection{Model Training}

We use ResNet-18 with Monte Carlo (MC) dropout. All models are initialized from ImageNet weights. Training details: Cross-Entropy loss; Adam optimizer with cosine warm-up; up to 50 epochs; early stopping (patience = 7, by validation loss); and AMP mixed precision. The best per-fold checkpoints (by validation performance) are retained for feature extraction and uncertainty estimation.

\subsection{Continuous Monitoring and Data Integration}

\subsubsection{Stage 1 (Feature Analysis \& Thresholds).}
Using each fold’s checkpoint, we extract $D$-dimensional penultimate-layer features for all base images, compute mean, variance/standard deviation, and covariance, and derive per-image Euclidean distance, cosine similarity, and Mahalanobis distance to the mean. Percentile gates are set as: 80th for Euclidean and Mahalanobis; 20th for cosine (higher is better).

\subsubsection{Stage 2 (Uncertainty \& Eligibility).}
\textbf{Test set:} For each image, 50 MC-dropout passes across 5 folds yield 250 predictions. We average softmax probabilities and compute predictive entropy. Relating entropy to correctness produces an ROC curve; maximizing Youden’s index yields an entropy threshold (e.g., $\sim0.247$) (\textbf{Figure 1}). 
\begin{figure}[t]
\centering
\includegraphics[width=0.85\linewidth]{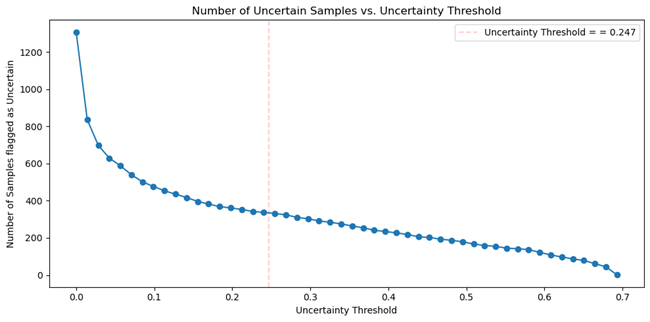}
\caption{\textbf{Stage-2:} Distribution of misclassified instances across different uncertainty values based on the ResNet18 model outputs on the test dataset. This figure highlights the relationship between predictive uncertainty and misclassification frequency, supporting the determination of an optimal uncertainty threshold.}
\label{fig1}
\end{figure}
\textbf{New images:} We apply the same 250-pass procedure, average features/probabilities, and require (i) distances/similarities within Stage-1 gates and (ii) entropy below the test-derived threshold. Only aligned, confident images are eligible.

\subsubsection{Stage 3 (Update \& Monitor).}
Eligible images with their predicted labels are added to the training folds. We retrain with identical hyperparameters and re-evaluate on the unchanged test set. Integration is accepted only if performance metrics (AUC, accuracy, sensitivity, specificity) do not degrade by more than 5\% and the entropy threshold does not increase by more than 5\%, computed as:
\[
\text{Percent Change} = \frac{(X - Y)}{Y} \times 100
\]

\section{Experiments and Results}

\setlength{\tabcolsep}{3pt}
\begin{table*}[h]
\centering
\caption{Distance and similarity metrics along with predictive uncertainty for five new images computed using an ensemble of 5-fold ResNet18 models. For each image, the Euclidean distance, cosine similarity, Mahalanobis distance, and predictive entropy are reported alongside their respective thresholds—80th percentile for Euclidean and Mahalanobis distances, 20th percentile for cosine similarity (both computed on the base dataset), and the predictive entropy threshold computed on the test dataset.}
\label{tab:resnet18_metrics}
\scriptsize
\begin{tabularx}{\textwidth}{c *{8}{>{\centering\arraybackslash}X}}
\toprule
\textbf{Image ID} &
\textbf{Euclidean Distance} &
\textbf{Euclidean Threshold} &
\textbf{Cosine Similarity} &
\textbf{Cosine Threshold} &
\textbf{Mahalanobis Distance} &
\textbf{Mahalanobis Threshold} &
\textbf{Predictive Uncertainty} &
\textbf{Predictive Uncertainty Threshold} \\
\midrule
6  & 13.29 & 18.47 & 0.8345 & 0.7476 & 18.81 & 25.10 & 0.00911 & 0.24682 \\
28 & 11.93 & 18.47 & 0.8371 & 0.7476 & 18.08 & 25.10 & 0.00197 & 0.24682 \\
57 & 11.99 & 18.47 & 0.8688 & 0.7476 & 17.36 & 25.10 & 0.11661 & 0.24682 \\
62 & 11.15 & 18.47 & 0.8454 & 0.7476 & 16.11 & 25.10 & 0.01079 & 0.24682 \\
70 & 13.99 & 18.47 & 0.8114 & 0.7476 & 21.19 & 25.10 & 0.00711 & 0.24682 \\
\bottomrule
\end{tabularx}
\end{table*}

\begin{table*}[h]
\centering
\caption{Performance metrics of the ensemble of 5-fold ResNet18 models (AUC for distinguishing proliferative from non-proliferative Glomerulonephritis) on the test dataset after retraining with the new image. Metrics such as AUC, accuracy, sensitivity, specificity, and the uncertainty threshold are reported along with the percent change from the original results, demonstrating that retraining results in negligible changes in model performance and uncertainty.}
\label{tab:resnet18_retrained}
\scriptsize
\begin{tabularx}{\textwidth}{c *{10}{>{\centering\arraybackslash}X}}
\toprule
\makecell{\textbf{Image}\\\textbf{ID}} &
\makecell{\textbf{Retrained}\\\textbf{AUC}} &
\makecell{\textbf{AUC}\\\textbf{\% Change}} &
\makecell{\textbf{Retrained}\\\textbf{Accuracy}} &
\makecell{\textbf{Accuracy}\\\textbf{\% Change}} &
\makecell{\textbf{Retrained}\\\textbf{Sensitivity}} &
\makecell{\textbf{Sensitivity}\\\textbf{\% Change}} &
\makecell{\textbf{Retrained}\\\textbf{Specificity}} &
\makecell{\textbf{Specificity}\\\textbf{\% Change}} &
\makecell{\textbf{Uncertainty}\\\textbf{Threshold}} &
\makecell{\textbf{Uncrtnty}\\\textbf{\% Change}} \\
\midrule
6  & 92.01\% & 0.20\%  & 88.98\% & -0.09\% & 59.92\% & -4.35\% & 96.10\% & 0.60\% & 0.032 & -87.08\% \\
28 & 92.06\% & 0.26\%  & 89.14\% & 0.09\%  & 61.09\% & -2.48\% & 96.00\% & 0.50\% & 0.165 & -33.10\% \\
57 & 92.03\% & 0.23\%  & 88.98\% & -0.09\% & 60.70\% & -3.11\% & 95.90\% & 0.40\% & 0.200 & -18.78\% \\
62 & 92.06\% & 0.26\%  & 89.06\% & 0.00\%  & 60.70\% & -3.11\% & 96.00\% & 0.50\% & 0.237 & -3.94\%  \\
70 & 91.99\% & 0.19\%  & 89.06\% & 0.00\%  & 60.31\% & -3.73\% & 96.10\% & 0.60\% & 0.024 & -90.14\% \\
\bottomrule
\end{tabularx}
\end{table*}

\subsection{Stage 1}

Per-fold feature summaries (mean, variance/std, covariance) define the reference distribution. The 80th/20th percentile gates operationalize in-distribution versus outlier detection (\textbf{Figure 2, Table 1}).

\begin{figure}[h]
\centering
\includegraphics[width=0.85\linewidth]{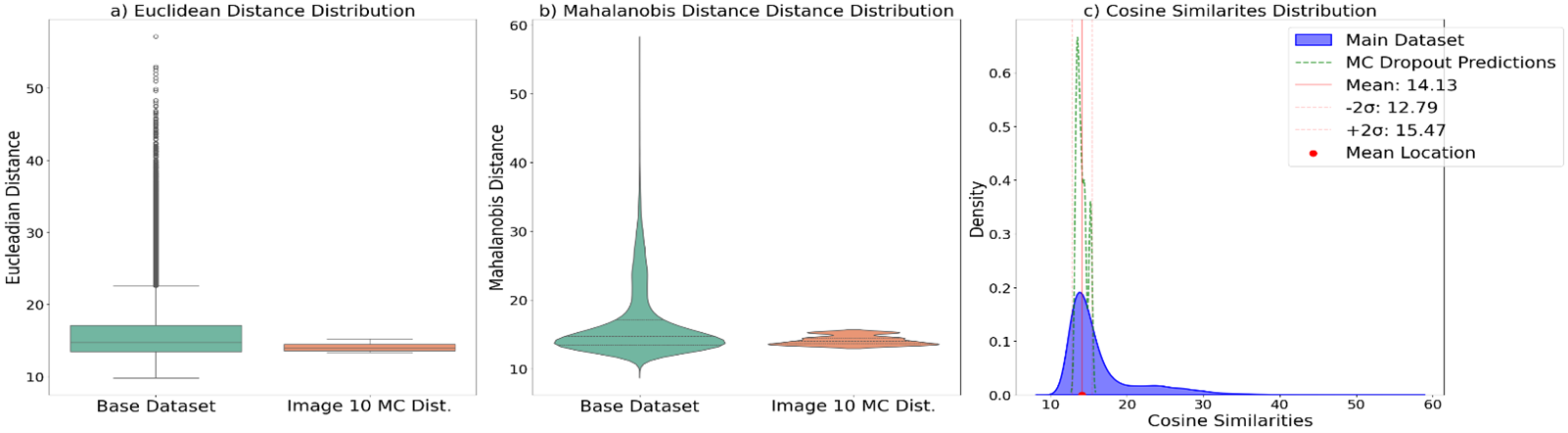}
\caption{\textbf{Stage-1:} Distributions of Euclidean distance, cosine similarity, and Mahalanobis distance for new image (Image ID: 6) computed over 250 Monte Carlo dropout iterations, overlaid with the corresponding base dataset distributions. This visualization demonstrates how the new image aligns with the base dataset according to the selected metrics.}
\label{fig:exp3}
\end{figure}

\subsection{Stage 2}

\textbf{Test set:} 250 predictions per image yield an ROC AUC of $\sim0.84$ for predicting misclassification via entropy, supporting an entropy threshold near 0.247.  
\textbf{New images:} Five illustrative cases satisfied all Stage-1 gates (low Euclidean/Mahalanobis, high cosine) and had mean entropies below $\sim0.247$ (\textbf{Table 1}).

\subsection{Stage 3}

Retraining each fold with a single eligible image produced negligible changes on the test set. AUC, accuracy, sensitivity, and specificity remained within $\pm5\%$, and the entropy threshold did not rise, indicating stable integration (\textbf{Figure 3, Table 2}).

\begin{figure}[htp]
\centering
\includegraphics[width=0.85\linewidth]{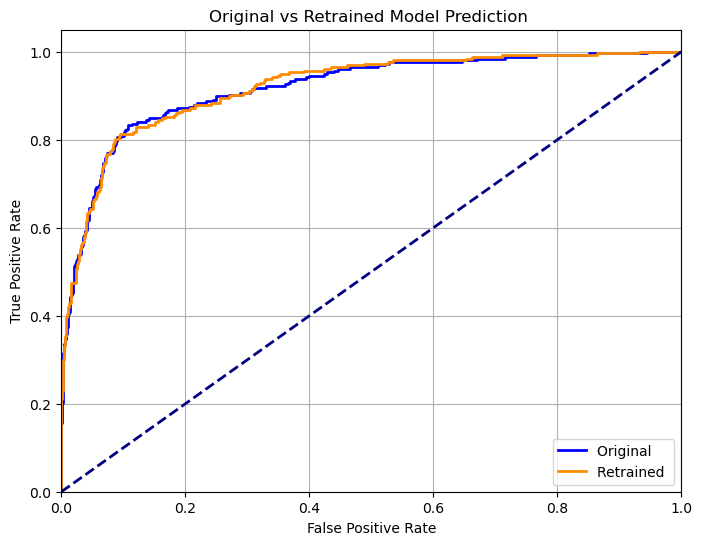}
\caption{\textbf{Stage-3:} Comparison of the ensemble of 5-fold ResNet18 models AUC for distinguishing proliferative from non-proliferative Glomerulonephritis on the test dataset before and after retraining with the new image. The negligible difference in the AUC values confirms that the retraining process maintains model performance while integrating the new image.
}
\label{fig:exp3}
\end{figure}

\section{Discussion}
We introduced a continuous learning framework that enables a medical imaging AI model to adapt to new data while preserving its existing accuracy. The combination of feature-space monitoring and uncertainty gating effectively addresses model degradation due to data drift. In our experiments, this approach prevented the kind of performance decay that can occur when a model encounters unseen variations. Unlike naive retraining (which could lead to catastrophic forgetting ~\cite{ref4}), our method ensured that each update was validated to be safe. This highlights a practical path toward deploying AI systems that evolve over time in a controlled manner. For instance, online model calibration methods have been explored to adjust predictions under shifting data ~\cite{ref8}, but our approach provides a different safeguard by using criteria to decide when retraining is necessary.
There are some limitations. Our threshold-based inclusion may reject unusual cases that fall outside the current distribution; incorporating a mechanism to eventually learn from such outliers (perhaps with human review) could further improve the system. We also added one image at a time – future work should investigate batch updates and dynamic threshold adjustment as data accumulates. Nonetheless, our findings suggest that rigorous continuous monitoring can make AI models more resilient. By selectively learning from new in-distribution, high-confidence examples, the model avoided both drift and forgetting, a key improvement over prior continuous learning approaches. We believe this strategy can generalize to other medical imaging tasks and help close the gap between static model performance in research and the need for ongoing reliability in clinical practice.

\section{Compliance with Ethical Standards}
This study was performed in line with the principles of the Declaration of Helsinki. The retrospective use of de-identified human subject data was approved by the institutional review boards (IRBs) of the University of Houston, University Hospital Cologne, Stanford University, and the University of Chicago. All data were anonymized prior to analysis.

\section{Acknowledgments}
\label{sec:acknowledgments}
This work was supported by NIH R01DK134055.

Dr. Mohan has consultancy or sponsored research agreements or equity with Boehringer-Ingelheim, Progentec Diagnostics, and Voyager Therapeutics. Dr. Mohan is on the Medical Scientific Advisory Council of the Lupus Foundation of America.
Dr. Mohan’s research is supported by NIH RO1 AR074096 and DK134055.

\bibliographystyle{IEEEbib}
\bibliography{main}

\end{document}